# Inverse design of photonic surfaces on Inconel via multi-fidelity machine learning ensemble framework and high throughput femtosecond laser processing

Luka Grbčić[1, †], Minok Park[2, †], Mahmoud Elzouka[2], Ravi Prasher[2, 3], Juliane Müller[4], Costas P. Grigoropoulos[2, 3], Sean D. Lubner[2, 5, *], Vassilia Zorba[2, 5, *], and Wibe Albert de Jong[1, *]

[1]*Applied Mathematics and Computational Research Division, Computing Science Area, Lawrence Berkeley National Laboratory, Berkeley, California, 94720, USA*

[2]*Energy Storage and Distributed Resources Division, Energy Technologies Area, Lawrence Berkeley National Laboratory, Berkeley, California, 94720, USA*

[3]*Department of Mechanical Engineering, University of California at Berkeley, Berkeley, California, 94709, USA*

[4]*Computational Science Center, National Renewable Energy Laboratory, Golden, Colorado, 80401, USA*

[5]*Department of Mechanical Engineering, Division of Materials Science and Engineering, Boston University, Boston, Massachusetts, 02215, USA*

[*]Corresponding author(s). E-mail(s): slubner@bu.edu; vzorba@lbl.gov; wadejong@lbl.gov
[†]These authors contributed equally to this work.

**Abstract**

We demonstrate a multi-fidelity (MF) machine learning ensemble framework for the inverse design of photonic surfaces, trained on a dataset of 11,759 samples that we fabricate using high throughput femtosecond laser processing. The MF ensemble combines an initial low fidelity model for generating design solutions, with a high fidelity model that refines these solutions through local optimization. The combined MF ensemble can generate multiple disparate sets of laser-processing parameters that can each produce the same target input spectral emissivity with high accuracy (root mean squared errors < 2%). SHapley Additive exPlanations analysis shows transparent model interpretability of the complex relationship between laser parameters and spectral emissivity. Finally, the MF ensemble is experimentally validated by fabricating and evaluating photonic surface designs that it generates for improved efficiency energy harvesting devices. Our approach provides a powerful tool for advancing the inverse design of photonic surfaces in energy harvesting applications.

## 1. Introduction

Modern photonic surfaces with tailored optical properties have been employed across various energy harvesting and storage applications, including thermophotovoltaics (TPV)[1,2], passive radiative cooling[3,4], solar water desalination[5,6], and concentrated solar power systems[7,8]. Such photonic systems allow manipulation of both light absorption and thermal emission, and may under certain conditions yield optical properties not frequently found in natural materials. Optical properties are determined by a surface's spectral absorptivity and emissivity at thermal equilibrium, which represent the energy absorbed or radiated from the actual surface at each wavelength normalized to that of a theoretically ideal surface[9,10]. Consequently, the ability to spectrally engineer emissivity translates to enhancement of the performance of these systems through radiative energy transfer control.

Recent advancements in machine learning (ML) approaches have enabled the inverse design of such photonic surfaces[11,12], including adversarial autoencoders[13], generative adversarial networks[14,15], and variational autoencoders[16]. Fully trained inverse ML models directly suggest design parameters to achieve desired optical properties, bypassing the need for iterative electrodynamics simulations. Nevertheless, previous efforts often overlook practical uncertainties and experimental validations, rely on computationally expensive algorithms, and provide a single design output. Their effectiveness is further limited by their ability to handle complex one design to many solutions mapping scenarios between design inputs and outputs when an inverse relationship is considered, which are common in real world manufacturing[17].

Previously, leveraging a combination of low-fidelity (LF) and high-fidelity (HF) ML models in optimization processes has showcased the strengths of combining both approaches, enhancing efficiency and outcome accuracy[18-20]. Starting with LF models, which are less computationally demanding albeit less precise, enables a rapid and broad exploration of parameter spaces. This initial phase narrows down areas of interest for subsequent analysis. Transitioning to HF models, which are accurate but computationally intensive, focuses resources on the promising regions identified by the LF models[21]. This strategic approach balances total computational cost and time with a target level of prediction accuracy and design diversity. The initial LF predictions' warm start circumvents exhaustive HF analysis across the entire parameter space, leading to more informed and precise outcomes with optimized resource allocation. Accordingly, we anticipate that a multi-fidelity (MF) ensemble framework integrating LF and HF models could (*i*) be lightweight and easily trainable, and (*ii*) generate multiple solutions with high accuracy to achieve target optical properties by leveraging complex one-to-many mappings, thereby advancing the inverse design of photonic surfaces and the associated ML-driven fabrication processes.

Pulsed laser ablation, a process involving material removal from a surface during laser material interactions, has been widely used to precisely alter and enhance surface properties on target materials[22-24]. Of particular interest is ultrafast femtosecond (fs) laser, which offers the ability to create a wide range of surface morphologies, spanning from the nanometer to hundreds of micrometer scale either via self-organization or direct laser writing[25,26]. Such surface structures can change the optical properties of a surface across the visible to infrared wavelength ranges on metallic substrates due to surface plasmon absorption or oxide formation mechanisms[27-29]. More importantly, such changes in spectral emissivity can be manipulated by adjusting laser parameters, involving power, scanning speed, and spacing between consecutive scan lines. However, due to the complex and multifaceted nature of the fs laser ablation process, modeling that considers laser parameters, fabricated surface morphologies, and resulting optical properties proves challenging[22,23]. Employing an ML approach capable of elucidating the direct mapping function from target optical properties to laser parameters and unraveling these intricate relationships holds significant potential in enabling the inverse design of photonic surfaces.

Here, we demonstrate inverse design of photonic surfaces using high throughput fs laser processing and the MF ensemble framework. Specifically, 11,759 photonic surfaces on Inconel are fabricated and optically characterized using fs laser processing and a custom microscope Fourier Transform Infrared spectrometer (FTIR). The MF ensemble, comprising an initial LF prediction phase followed by a refined HF optimization model, is developed and trained on experimentally obtained data. This model offers ease of training, and produces multiple input parameters not present within the training dataset, as well as design outputs to achieve target optical properties by utilizing the complex one-to-many mapping relationship. SHapley Additive exPlanations (SHAP) is used to verify laser parameters that most significantly influence optical properties. Lastly, we show the capability of the trained MF ensemble to achieve inverse design of photonic surfaces with different optical properties on demand. Our approach, which integrates fs laser processing and MF ensemble, underscores the ability to facilitate the inverse design of photonic surfaces for energy harvesting applications.

## 2. High throughput fs laser fabrication and optical property characterization

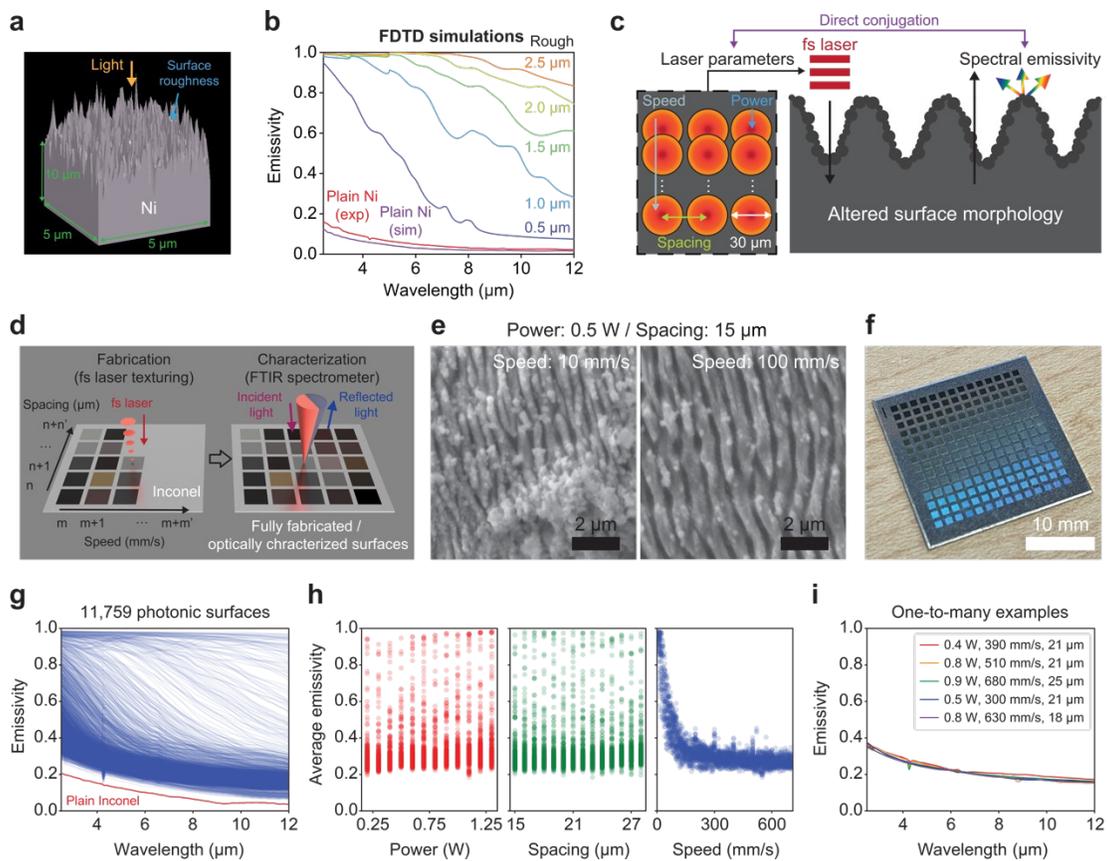

**Figure 1.** High throughput fs laser fabrication and optical property characterization of photonic surfaces on Inconel for training data. (a) Schematic illustration of FDTD simulations, and (b) the simulated spectral emissivity for Ni substrates with different surface roughness. (c) Schematic of fs laser processing by controlling three independent parameters (laser scanning speed, spacing, and power), which control processed surface morphology and spectral emissivity. The focused laser spot diameter is 30 µm. (d) Schematic representation of automated high throughput fs laser

fabrication and optical property characterization using FTIR. (e) SEM images showcasing two representative surface morphologies, fabricated using the same laser power of 0.5 W and laser spacing of 14 μm, but different scanning speeds of 10 mm/s (left image) and 100 mm/s (right image). Black scale bars are 2 μm. (f) Example photograph of fabricated 196-morphologies on an Inconel substrate. The white scale bar is 10 mm. (g) Spectral emissivity measurements for all 11,759 surface structures. (h) Distribution of unweighted average emissivity for all 11,759 structures as a function of laser power, spacing, and speed. (i) Example of five disparate sets of laser parameters that produce nearly the same spectral emissivity (one-to-many mapping).

During fs laser material interactions, the target material undergoes melting and evaporation when the deposited laser energy raises its temperature above the melting point[22-24,30]. This leads to ablation dynamics which involve the expulsion and subsequent solidification of the surface material, yielding diverse surface morphologies including nanoparticles and microstructures.

To examine the impact of surface morphologies on optical properties, finite difference time domain (FDTD) simulations using Lumerical (Ansys Inc.) are employed on Ni substrates with varying surface roughnesses, mimicking nanoparticles, as shown in Figure 1a. Ni is selected because it is the major component for Ni-based superalloys (e.g., Inconel), which are widely employed in high-temperature applications[31]. As presented in Figure 1b, the spectral emissivity increases with higher surface roughness, confirming different morphologies can produce diverse spectral emissivities that deviate from pristine optical properties in infrared (IR) wavelengths. Such morphological variations can be achieved by factors including laser power, scanning speed, and spacing (Figure 1c). Consequently, establishing a direct mapping between laser parameters and spectral emissivity through ML models can expedite the achievement of target optical properties without unveiling interactions between surface geometries and optical properties.

As outlined in Figure 1d, we employ a high throughput fs laser fabrication to build datasets (Supplementary Figure 1a). Specifically, an ultrafast fs laser (500 fs pulse duration, 1030 nm wavelength, and 30 μm focused beam diameter) is utilized to fabricate a variety of surface geometries. Each surface exhibits different spectral emissivity contingent upon distinct laser processing conditions. For example, modifying only the speed while maintaining two variables constant (0.5 W power and 15 μm spacing) results in more pronounced surface structures at a speed of 10 mm/s compared to 100 mm/s, as shown in scanning electron microscopy (SEM) images (Figure 1e). Varying laser power (0.2 W to 1.3 W in 0.1 W increments; see Supplementary Table 1 for laser conditions in intensity and fluence), scanning speed (10 mm/s to 700 mm/s in 10 mm/s increment), and line spacing (15 μm to 28 μm in 1 μm increments) produce a total of 11,759 combinations (Supplementary Figure 1b). As shown in Figure 1f, each combination within these three-dimensional parameters is applied to every 1 mm$^2$ area on Inconel, with each application requiring less than a couple of seconds. This process facilitates high throughput fabrication and associated data generation.

Following the fabrication process, spectral emissivities of all 11,759 fabricated surfaces are characterized within the wavelength range of 2.5 μm to 12 μm using a custom microscope FTIR optical property characterization system, as shown in Figure 1g and S2. Average emissivities of all samples are shown in Figure 1h and Supplementary Figure 3, ranging from 0.2 to 1.0 as a function of three laser parameters. While the impact of laser power and spacing on optical property variation is less pronounced, scanning speed primarily affects average emissivity, with lower speed resulting in higher spectral emissivity. This indicates that a lower scanning speed allows for more pulses to modify the target surface, deviating from the original flat surface (e.g., Figure 1e). Conversely, higher speed yields lower average emissivities, approximating the pristine substrate's average emissivity of 0.14, due to the utilization of fewer laser pulses. Figure 1i shows an example of the one-to-many mapping scenarios between laser parameters and spectral emissivity, demonstrating that multiple laser parameter sets can lead to similar optical properties.

**3. Architecture of the MF ensemble framework**

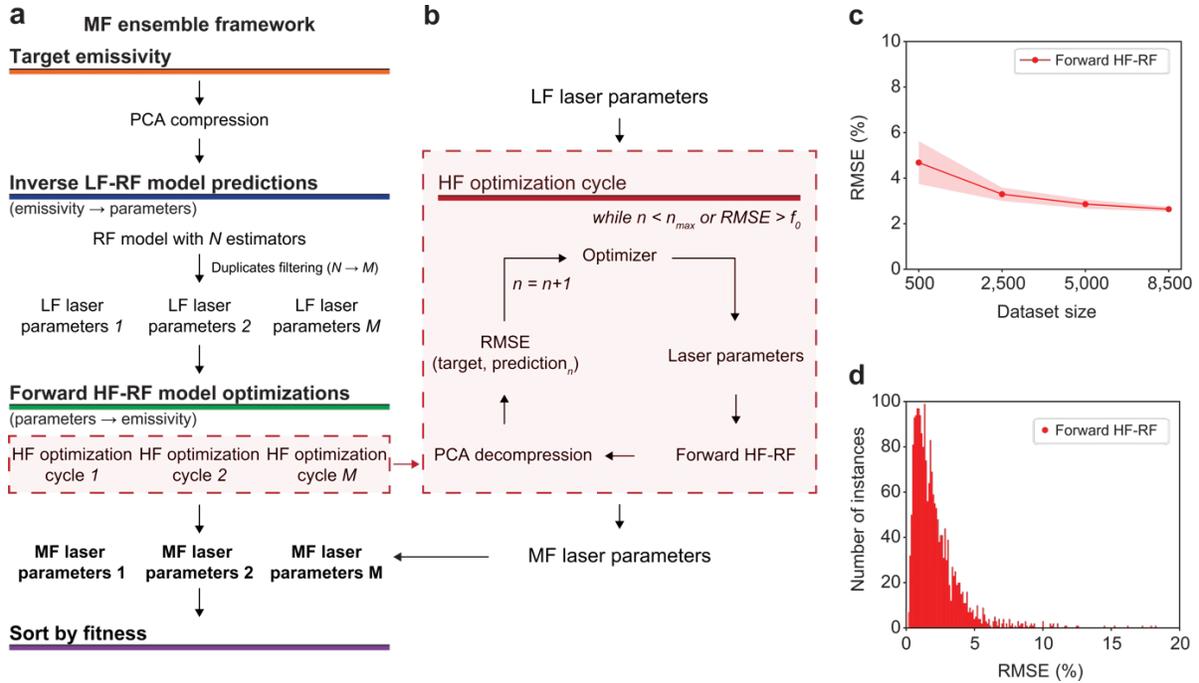

**Figure 2.** MF ensemble framework. (a) MF ensemble framework architecture. For each individual target spectral emissivity, the inverse LF-RF model generates M unique laser parameter sets from anywhere in the full global parameter space. These laser parameter predictions are then refined through local optimization using the forward HF-RF model (separately trained RF). (b) Individual steps in the HF-RF local optimization cycles, which refine the laser parameter values to minimize the difference between their resulting spectral emissivity (as predicted by the forward HF-RF model) and the target spectral emissivity. The HF optimization cycle is terminated either when the maximum number of evaluations $n_{max}$ is reached or when the obtained RMSE value is less than the desired fitness threshold, $f_0$. Forward HF-RF model results: (c) learning curve with K-Fold cross-validation with $K = 10$ and RMSE. (d) The RMSE distribution of the forward HF-RF model predictions, where the train and test data were obtained with a separate 75%/25% random split of the 8,500-training dataset.

The MF ensemble framework (Figure 2) is specialized for the inverse design task of systems with one-to-many mappings, which appears in this work as each spectral emissivity curve corresponding to multiple distinct sets of laser manufacturing parameters. An LF inverse model first generates multiple approximate design solutions from the full global design parameter space for a given target input property. An optimization algorithm then uses an HF forward model to locally refine each of these designs to minimize the loss between their resulting properties and the target property. In this work, we use Random Forests (RF) for both the LF and HF models (thus denoted as LF-RF and HF-RF, respectively), and a Differential Evolution (DE) algorithm for our optimizer. However, the general MF ensemble framework is agnostic to the specific models chosen and offers advantages including a computationally lightweight and modular structure where we can easily swap out different models. It can effectively handle various inverse design tasks[32-34], and warm starting the HF optimizations using the LF predictions yields benefits such as improved convergence rates given a limited computational budget and fewer hyperparameters compared to deep learning models[35].

The training process commences with a target emissivity curve compressed using the Principal Component Analysis (PCA) algorithm (details in the Methods and Supplementary Figure 4). Next, the inverse LF-RF model generates $N$ sets of laser parameters based on this PCA compressed emissivity. Duplicates are filtered out, leaving $M \leq N$ unique sets. The DE algorithm uses the HF-RF model to iteratively optimize each design to identify $M$ optimal laser parameter sets that each minimize the root mean squared error (RMSE) between the target and their predicted emissivities (Figure 2b). The $M$ designs are ranked according to their calculated fitness metric, $f$, which in this work is RMSE. The optimization process is terminated upon meeting either of two criteria: reaching the maximum number of evaluations ($n_{max}$), or reducing RMSE below a fixed fitness threshold, $f_0$.

To train the forward HF-RF model within the MF ensemble architecture, the initial experimental dataset is randomly split into a train/validation set (8,500 samples) and a test set (3,259 samples), as shown in Supplementary Figure 5. The train/validation set is utilized to assess the robustness and accuracy of the HF-RF model (i.e., standard deviation and average of RMSE), while the test set evaluates the performance of the ML ensemble (Supplementary Figure 6). As presented in Figure 2c, the forward HF-RF model demonstrates the highest accuracy (an average RMSE of 2.6%) when trained with the largest dataset size in a cross-validation strategy (detailed in Equation 2). Furthermore, the majority of the predicted instances show RMSEs below 5% when using a 75/25% train/test split (out of the 8,500 samples), indicating the HF-RF model's ability to accurately predict emissivity values and integrate seamlessly into the ML ensemble (Figure 2d).

Training standalone or single-fidelity ML algorithms for single instance inverse design presents challenges due to the complex one-to-many mapping scenarios (e.g., Figure 1i), making it difficult to effectively minimize the loss function. Therefore, as shown in Supplementary Figure 7, we conduct comprehensive testing using three different hyperparameter-tuned inverse ML models: standalone RF, Light Gradient Boosting Machine-LightGBM (LGB), and eXtreme Gradient Boosting (XGB) algorithms, known for their suitability with well-structured tabulated features. Notably, each standalone model yields insufficient accuracy with RMSE higher than 10%, in mapping spectral emissivity to laser processing parameters. Due to the demonstrated inaccuracy of the standalone inverse ML models, the MF ensemble comprising the inverse LF-RF and forward HF-RF models proves particularly advantageous in achieving accurate predictions when dealing with one-to-many mapping between design inputs and outputs. Moreover, owing to its capability to generate several predictions from a single input sample, the RF algorithm is a critical component of the MF ensemble. It facilitates the initiation of numerous HF optimization cycles through a warm-starting process. Additionally, given its demonstrated high accuracy in modeling the forward relationship, as shown in Figure 2c and 2d, the RF algorithm is an excellent choice for the forward model in the HF optimization cycle.

**4. Performance of the MF ensemble framework**

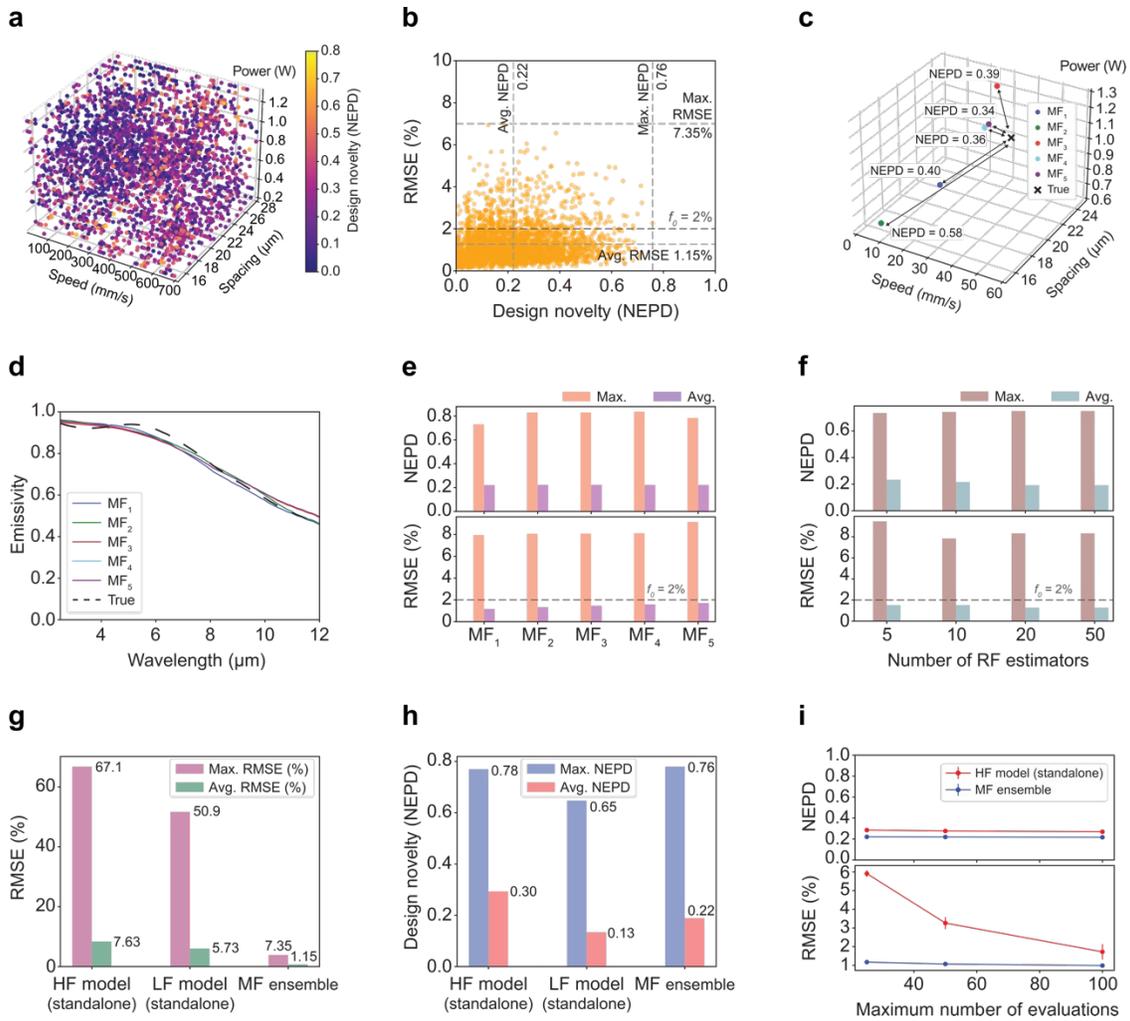

**Figure 3.** Performance of the trained MF ensemble on the test set ($N$ = 3,259). (a) The MF ensemble-predicted laser parameters for the test set ($N$ = 3,259), colored according to their NEPD value. (b) Corresponding NEPD plot versus RMSE of the ML ensemble where the vertical lines show the average and maximum NEPD, while the horizontal lines show the average and maximum RMSE. (c) An example of 5 sets of predicted laser processing parameters for the same target emissivity curve, shown in the parameter design space alongside the true set of laser processing parameters for that emissivity curve. (d) The forward HF-RF model-predicted spectral emissivity curves (all below 3% RMSE) of the predicted laser processing parameters shown in (c), juxtaposed over the original target spectral emissivity ("True"). (e) Average and maximum NEPD and RMSE for the top five predicted designs for each target emissivity curve in the test set, grouped according to fitness ranking. (f) Average and maximum NEPD and RMSE for the top-ranked predicted design for each target emissivity curve in the test set, when the number ($N$) of inverse LF-RF estimators is varied. (g) RMSE comparison among three inverse design models: HF model standalone (no LF model warm start), randomly initialized with a fixed number of 25 evaluations, the LF model standalone (no HF optimization), and the full MF ensemble with a fixed number of 25 HF evaluations. (h) NEPD comparison among the same three models in (g). (i) RMSE and NEPD for the MF and standalone HF models for varying fixed numbers of evaluations, $n$, of each. Points and error bars for NEPD of the predicted laser parameters and RMSE of the predicted spectral emissivities are calculated from a 5 run average of the test set.

Figure 3 shows the performance evaluation of the fully trained MF ensemble. Design novelty in laser parameters is assessed using normalized Euclidean parameters distance (NEPD) metrics (Equations 4, 5, and 6), which quantify the normalized deviation from the original laser parameters of the test set. An NEPD value of 0 indicates identical laser input parameters while 1 implies the maximum possible difference between two sets of laser parameters. The predicted laser parameters obtained from the test set's emissivity (Supplementary Figure 6a) are compared to experimental laser parameters (Supplementary Figure 6b) using NEPD. These predicted parameters are input into the pre-trained forward HF-RF model to predict spectral emissivity, which is further compared with the original test set spectral emissivity using RMSE, as shown in Supplementary Figure 8.

Figure 3a shows the top-ranked design predicted by the trained MF ensemble for each spectral emissivity curve in the test set ($N$ = 3,259, Supplementary Figure 6a), with color denoting NEPD value. These parameters show a uniform scatter across the entire parameter space, indicating the model's capacity to generalize and explore diverse solutions without bias toward confined laser parameter spaces. Figure 3b shows the predicted designs' RMSE and NEPD values. The average and maximum NEPD (0.22 and 0.76, respectively) shows that the model frequently predicts meaningfully novel designs, utilizing the one-to-many mapping between the spaces. Additionally, the degree of novelty does not correlate with RMSE, suggesting the model generalizes robustly without loss of accuracy. The low average and maximum RMSE (1.15% and 7.35%, respectively) suggest that the majority of optical properties across the full properties space can be accurately approximated. Figure 3c and 3d collectively illustrate the analysis of emissivity curves and laser parameters. Figure 3c presents a diverse set of the top five fitness-ranked predictions for one target emissivity curve, demonstrating variability in the solutions. In contrast, Figure 3d shows the reconstructed spectral emissivities by the forward HF-RF mode, which consistently match the original target emissivity curve ("True"). All five predicted emissivity curves closely match the target (average RMSE < 2%), while their corresponding designs are highly distinct in laser parameter space, reflecting high NEPD values. This example demonstrates how the MF framework successfully handles the one-to-many mapping, producing multiple accurate and novel designs for a single target. Figure 3e shows that on average the top five ranked generated designs perform similarly on RMSE and NEPD for any given target emissivity curve, demonstrating that the example in Figure 3c and 3d is representative of the full test set and hence full emissivity property space. Additional fitness-sorted prediction sets and their distributions are shown in Supplementary Figure 9, with an overlap and Gaussian density plot (Supplementary Figure 10). Moreover, despite the MF ensemble being executed with the $n_{max}$ set at 25, $f_0$ at 2%, and the number of estimators $N$ at 20 for the analysis of the top five ranked predictions, the RMSE and NEPD values remain nearly constant across all values of $N$ (Figure 3f), indicating that the hyperparameter which determines $N$ does not significantly affect the MF ensemble's performance.

To further put the effectiveness of the MF ensemble architecture in perspective, we characterize the RMSE and compare it with the standalone HF and LF models, as shown in Figure 3g and Supplementary Figure 11. The standalone HF model employs the same optimization cycle as in the MF model, except it is initialized with a set of random laser parameters instead of a laser parameter set predicted by the LF-RF inverse model for the particular target emissivity. The standalone LF model employs the same LF-RF model as in the MF model, but without subsequent local optimization of those predicted designs. It solely relies on the inversely predicted laser parameter sets as the final predictions, which are then averaged to provide a final prediction. The MF ensemble exhibits an average and maximum RMSE of 1.15% and 7.35% with prediction data (inference time presented in Table S2), outperforming both standalone HF and LF models by around a factor of five or more. The standalone HF model demonstrates an average RMSE of 7.63%, slightly less accurate than the standalone LF model (average RMSE of 5.73%). It is notable that the combination of the LF warm start with HF optimization significantly outperforms either in isolation. This behavior implies that for the nonlinear one-to-many mapping of these optical systems, the inverse LF-RF model on its own is poor at finding local minima, while the DE optimization with HF-RF on its own is poor at finding regions of global minima. Furthermore, the high accuracy of the MF ensemble is not achieved at the expense of its ability to attain high NEPD. Both the average and maximum NEPD for the MF ensemble are comparable to those of the standalone LF and HF models (Figure 3h).

A detailed comparison between the standalone HF model and the MF ensemble, considering various $n_{max}$ values with uncertainty analysis (5 repeated measurements per $n_{max}$ setting), is shown in Figure 3i. The NEPD is approximately independent of varying $n_{max}$ values because iterative HF evaluations only produce small local refinements in a prediction's location within the parameter space, and so do not significantly impact NEPD. However, high NEPD values for the standalone HF model could be due to inaccuracies in laser parameter predictions, as indicated by RMSE values, and, to a lesser degree, due to the broader exploration of the laser parameter space, given the random initialization of optimization cycles. The HF model's average NEPD marginally decreases as prediction accuracy improves (from 0.29 at $n_{max}$ = 25 to 0.27 at $n_{max}$ = 100). Considered together, Figures 3g and 3i indicate that a primary benefit of the HF model in the MF ensemble is to mitigate outliers among the LF model's initial predictions. The maximum error of any MF ensemble prediction across the entire test set was 7.4%, whereas outliers in the standalone models exhibited RMSE > 50%, which is far more likely to be detrimental in a real application. The MF ensemble's average RMSE at $n_{max}$ = 25 is 1.17%, compared to the standalone HF model's 1.72% at $n_{max}$ = 100. With sufficient $n_{max}$ evaluations facilitating further exploration of the laser parameter space, the standalone HF model could achieve accuracy comparable to the MF ensemble, as indicated by the declining RMSE trend. Nevertheless, significant RMSE uncertainty persists at $n_{max}$ = 100 due to the standalone HF model's random initialization and potential for getting trapped in local optima. Conversely, the MF ensemble, leveraging initial inverse LF-RF predictions, exhibits minimal prediction uncertainty, even at the lowest $n_{max}$ of 25. This comparison highlights the benefit of using the MF ensemble as it greatly balances the exploitation and exploration needed for this inverse design task, even for lower values of $n_{max}$.

**5. SHAP features importance analysis of the forward HF-RF model**

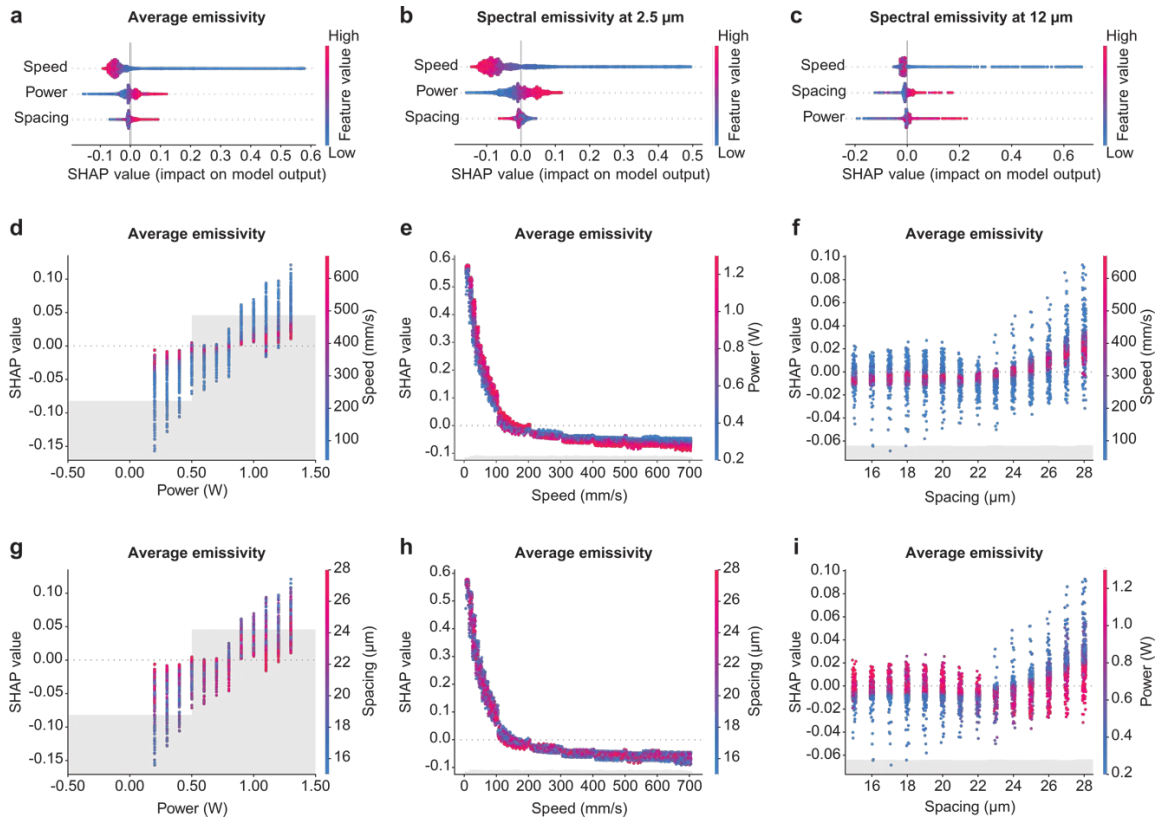

**Figure 4.** SHAP analysis of the forward HF-RF model. Laser processing parameters SHAP analysis on (a) average emissivity, (b) spectral emissivity at 2.5 μm wavelength, and (c) spectral emissivity at 12 μm wavelength, as the output values. The relationship for the average emissivity between the SHAP value and (d) the power colored by the speed feature, (e) the speed colored by the power feature, (f) the spacing colored by the speed feature, (g) the power colored by the spacing feature, (h) the speed colored by the spacing feature, and (i) the spacing colored by the power feature.

The utilization of the MF ensemble offers an advantage in that tools like SHAP[36,37] can be easily applied for segments of the inverse design framework, providing deeper insights into how different input features influence the model's predictions and reducing their black box nature. This is in stark contrast to traditional models that heavily rely on latent features and simply assign weight coefficients, which often require complex interpretation.

Figure 4a-c presents the results of the SHAP analysis, showcasing the impact of individual laser parameters on average emissivity and spectral emissivity at 2.5 and 12 μm wavelengths. Here, the SHAP quantifies the influence of each laser parameter on the deviation from the model's

average predicted output value (e.g., a SHAP of 0 in Figure 4a indicates the mean value of predicted average emissivities for the 8,500 datasets). The input features are ranked based on their influence on the baseline value, with speed being the top feature. In Figure 4a, lower speed values (depicted in blue) correspond to a positive SHAP value, suggesting an increase in average emissivity, consistent with the experimental observation shown in Figure 1h. However, higher values of power and spacing (represented in red) exhibit a positive correlation with average emissivity. In particular, the emissivity at a wavelength of 12 µm is more influenced by laser parameters (Figure 4b-c) than at 2.5 µm. Lower speed, higher power, and larger spacing contribute to increased emissivity at 12 µm wavelength. However, there is a reversal in the relationship between spacing and spectral emissivity between 2.5 µm and 12 µm wavelengths.

Figure 4d-i presents the individual parameter contributions of each feature to the SHAP for the average emissivity model (note that the *y*-axis scales differ). Increasing laser power tends to modestly increase average emissivity (Figure 4d and 4g), but at high scanning speeds the power has almost no influence (i.e., SHAP close to 0, Figure 4d). Conversely, decreasing scanning speeds tends to increase average emissivity regardless of power or spacing (Figure 4e and 4h), and this effect is especially strong for speeds below approximately 100 mm/s.  The spacing between laser scan lines does not strongly influence average emissivity either way (Figure 4f and 4i) and is especially neutral at high scan speeds (Figure 4f) akin to the observations of Figure 4d. But for scan line spacings larger than 22 µm there is a reversal in the slight influence of laser power on average emissivity (Figure 4i). Plots showing the individual contributions of each feature to the SHAP for the 2.5, 7.25 and 12 µm wavelength emissivity models are presented in Supplementary Figure 12, 13, and 14, respectively.

**6. Inverse design of photonic surfaces via the MF ensemble**

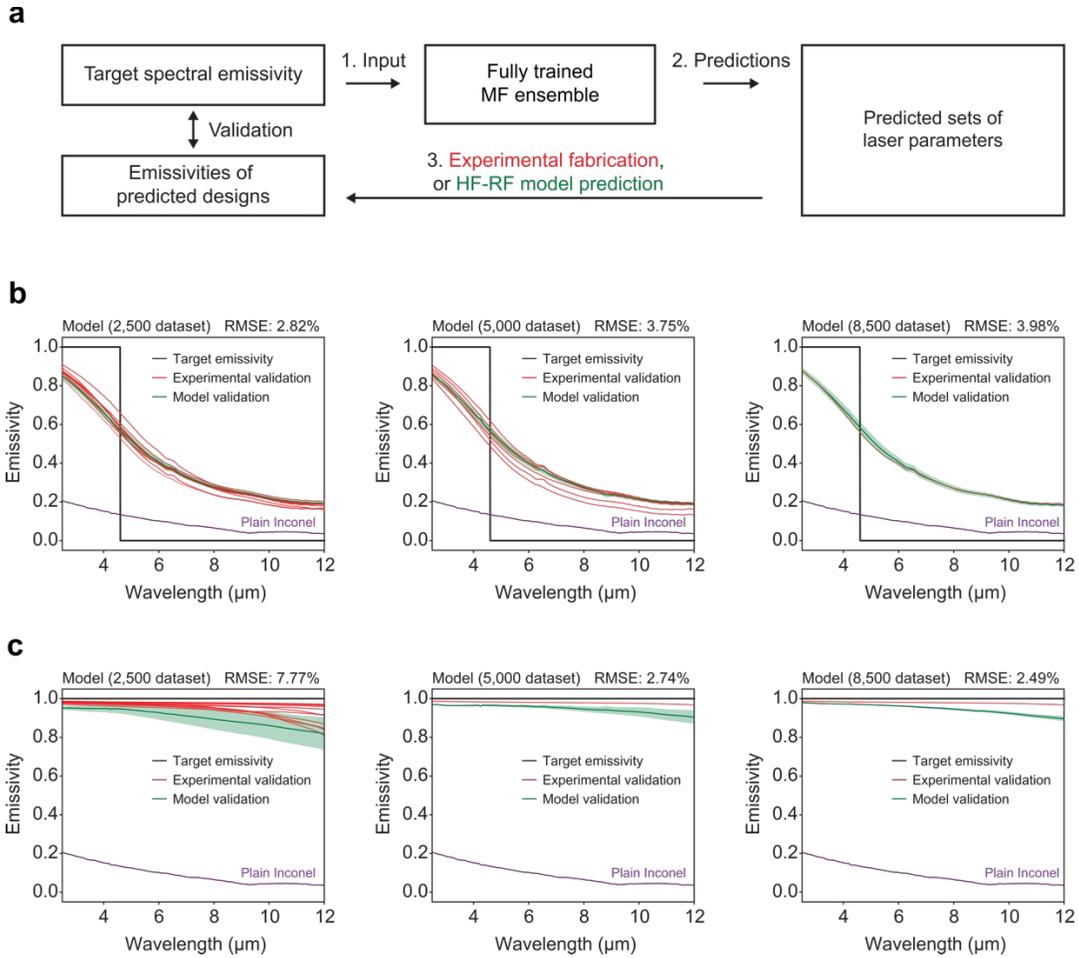

**Figure 5.** MF ensemble validation via inverse design of photonic surfaces. (a) Schematic representation of the validation process. The same fully trained MF ensemble predicts multiple sets of laser parameters for each target spectral emissivity. A subset of the design predictions is experimentally fabricated and their emissivity is measured using an FTIR spectrometer. The emissivities of the remaining designs are predicted using the HF-RF forward model. Both sets of emissivities are compared to their original design target for validation, quantified using RMSE. In the MF ensemble, the $n_{max}$ is set at 25, with a fitness threshold $f_0$ of 2%, and the number of top-ranking predicted laser parameter sets is set to 10, with the number of estimators $N$ set at 20. Inverse design of photonic surfaces targeting (b) selective TPV thermal emitter with a step at a wavelength of 4.6 μm and (c) near-perfect emitter, using the MF ensemble trained by 2,500, 5,000, and 8,500 data.

Finally, we demonstrate the utility of the MF ensemble by applying it to the inverse design of photonic surfaces for energy applications. Tailored surface optical properties are essential to performance and efficiency in a variety of energy harvesting and storage applications that rely on the radiative transport of energy. As schematically shown in Figure 5a, we use the same fully trained MF ensemble (properties presented in Figure 2, 3, and 4) to predict multiple sets of laser manufacturing parameters to produce the desired optical properties for specific applications, and then experimentally fabricate and validate these predicted designs. The validation of their optical properties is conducted either by confirming their presence in the original experimental dataset not used for MF ensemble training—or through the HF-RF model. We consider two applications: a spectrally selective thermal emitter for a lead-selenide TPV system, and a near-perfect blackbody thermal emitter.

TPVs convert high temperature thermal radiation into electricity using photovoltaic cells[2,38], and a lead-selenide based TPV operating at a temperature of 1400K with a bandgap of 4.6 μm aligns with our dataset's spectral range (according to Wien's displacement law), as shown in Figure 5b. The corresponding ideal thermal emitter should exhibit a spectral emissivity of 1 for wavelengths shorter than the bandgap and 0 for wavelengths longer than the bandgap to co-maximize heat-to-electricity conversion efficiency with generated power density. An ideal blackbody emitter has an emissivity of 1 for all wavelengths, thereby maximizing radiative heat transfer ("Target emissivity" in Figure 5c). These distinct target emissivities (black curves) are used as inputs to obtain laser parameters predicted by three MF ensembles (each trained on datasets of 2,500, 5,000, and 8,500), and the emissivity curves are subsequently estimated by the forward HF-RF models trained by the same datasets. Note that the exact same MF ensembles are used to generate predictions for both types of emissivity targets (i.e., TPV emitter and blackbody emitter). Moreover, this target emissivity challenges MF ensembles because it is qualitatively different from all training data and is not physically achievable using Inconel.

Due to the stochastic nature of the entire MF ensemble (Figure 2), we conduct 100 repeated runs for each target emissivity using each model to generate the MF ensemble laser parameter sets. We extract the top 10 ranked solutions for each run, removing any duplicates and predicted laser parameter sets that are in the training set. This process yields two groups of predicted laser parameter sets: (*i*) sets that are experimentally validated by determining if they are in the part of the experimental datasets not used for MF ensemble training, and (*ii*) sets that are HF-RF model validated. The laser parameter sets from group (*i*) are individually visualized as red lines, while those from group (*ii*) are averaged, with the shaded green region representing the standard deviation (green lines). The predicted laser parameters are shown in Supplementary Figure 15, and the statistics for groups (*i*) and (*ii*) for the 100 runs for all dataset sizes and targets are tabulated in Supplementary Table 3.

For the lead-selenide TPV thermal emitter target (Figure 5b), all trained MF ensembles yield spectral emissivity curves resembling the target ideal emissivity. The spectral emissivity at a 2.5 μm wavelength (shorter than bandgap) exceeds 0.8, while approaching 0.2 at a 12 μm wavelength (longer than bandgap). All three MF ensembles (trained on 2,500, 5,000, or 8,500 data) generate similarly high accuracy designs (2.8% < RMSE < 4%). However, the optical properties of the design predictions by the 8,500 data-trained ensemble are more consistent and tightly clustered, as seen in Figure 5b. Similar trends are observed for the blackbody emitter target, as shown in Figure 5c. The majority of the predictions from all models provide broadband spectral emissivity values that satisfy the target with lowest determined RMSE values, with more consistent and tightly clustered predictions from models trained on more data. However, for this target, the 2,500 data model performs noticeably worse, indicating the challenge of generating high Inconel emissivities at longer wavelengths deeper into the infrared wavelength. These results experimentally validate the MF ensemble's ability to predict multiple and diverse novel sets of laser parameters for a single spectral emissivity target, including when the target qualitatively deviates from the training data.

## 7. Conclusion

We demonstrate that the integration of high throughput fs laser fabrication and optical property characterization techniques with the MF ensemble framework enables precise inverse design of photonic surfaces. The fully trained MF ensemble adequately solves the inverse design problem for this class of microtextured optical surfaces with a complex one-to-many mapping relationship between desired optical property inputs and design parameter outputs. For each single target input emissivity, the MF ensemble predicts multiple distinct manufacturing designs that are spread throughout the design parameter space and often were never seen during training. The predictions are validated over a wide range of spectral emissivities, including experimental validation for two energy technology applications each with target emissivities that are qualitatively different from the training data. While this study primarily focuses on introducing, training, and testing the MF ensemble using three laser parameters and a specific material, our approach can be extended to the inverse design of photonic surfaces that may require different materials, additional laser parameters, or other relevant design parameters. Optimizing optical properties and understanding their complex functional relationship to target device design details are essential for boosting system efficiency and performance. This adaptability generalizes applicability to a broad diversity of possible energy harvesting and storage applications, such as heliostats, parabolic troughs, solar-water desalination, and passive radiative cooling. Furthermore, our approach can be extended to other laser processing applications with complicated relationships between laser parameters and materials' properties[22,23].

**Methods**

**Materials**

Inconel 625 substrates (GoodfellowUSA) with 0.5 mm thickness are used as target specimens.

**High throughput fs laser fabrication**

A fs laser system (s-Pulse, Amplitude), operating at a wavelength of 1030 nm with a pulse duration of 500 fs and a 100 kHz repetition rate, is employed for this study. The laser beam is focused via a galvano scanner (excelliSCAN 14, SCANLAB) with a beam spot size of 30 µm in diameter. This configuration enables the fabrication of diverse surface morphologies on Inconel under varying laser processing parameters on demand. A total of 11,759 surfaces (distinct 1×1 $mm^2$ areas) are fabricated using three laser parameters (power, speed, and spacing) with the raster scanning method, as shown in Supplementary Figure 1b. Each parameter combination is applied to 1 $mm^2$ areas of Inconel substrates.

**High throughput optical property characterization**

A custom microscope Fourier Transform Infrared spectrometer (Thermo Fisher Scientific, Nicolet iS50) microscope system is established for direct high throughput optical properties measurement of fabricated morphologies, as illustrated in Supplementary Figure 2. The system utilizes an optical microscope configuration with a reflective objective lens and a liquid nitrogen cooled Mercury-Cadmium-Telluride detector, enabling precise measurements of spectral reflectivity and corresponding emissivity within the wavelength range of 2.5 µm to 12 µm. The system is synchronized with a set of motorized *XYZ* stages for automated and high throughput measurements.

**Computational resources**

Computational resources used for ML training and analyses include a personal laptop computer (Lenovo Thinkpad X1 ExtremeGen 4 with 11th Gen Intel Core i7-11800H (2.30 GHz) and 16 GB RAM).

**MF ensemble framework architecture**

The goal of the end-to-end MF ensemble-informed inverse design model architecture is to extract multiple sets of solutions (i.e., a set of laser parameters) at real time that will accurately obtain the desired optical properties (i.e., target spectral emissivity). To achieve this, the model goes through two main stages that involve inverse LF-RF model prediction and forward HF-RF model-based optimization. Firstly, the target spectral emissivity is compressed with a pretrained PCA model to increase the computational efficiency by reducing the number of features. The chosen number of components in the PCA model is explained in the Data preprocessing section.

The initial stage of the model involves leveraging a pre-trained RF algorithm, which utilizes the compressed target emissivity curve as input to predict $N$ sets of laser parameters. More specifically, the RF algorithm constructs $N$ decision trees (DTs), where $N$ serves as a tunable hyperparameter. These trees are built using random feature subsets, and while their outputs are typically averaged, this application makes use of each DT's individual prediction. These predictions are part of a LF prediction stage, where pinpoint accuracy for each parameter is not critical.

Before the second stage of the model is started, the number of RF predictions $N$ are filtered for duplicates to obtain $M$ laser parameter sets. These $M$ laser parameter sets are then used as initial guesses for a HF stage of the inverse design model in order to increase the accuracy and computational efficiency of the spectral emissivity reconstruction. An optimization cycle is started for each laser parameter set separately, and the pre-trained forward HF-RF model maps laser parameters to spectral emissivity. Each optimization cycle uses an optimization algorithm to generate new laser processing parameter solutions which are evaluated by the forward HF-RF model and subsequently the decompressed predicted spectral emissivity is compared with the target spectral emissivity to assess the fitness of the generated laser parameters. The optimization goal function and boundaries (Equation 1) are defined in the Optimization goal function and algorithm section.

The HF optimization process is governed by two hyperparameters: the maximum number of evaluations ($n_{max}$) and the fitness threshold ($f_0$). These parameters dictate the termination of the optimization loop, which concludes either after reaching $n_{max}$ iterations or achieving a fitness level below the threshold $f_0$, indicated by a sufficiently low RMSE (defined in Equation 2) between the predicted and target spectral emissivities. The DE global stochastic optimization algorithm was used in the HF optimization cycle due to its suitability for non-linear and multimodal problems. Upon completion of the optimization process with the forward HF-RF model, the laser parameters are then sorted based on their respective fitness scores. This ranking facilitates the formation of solution sets for each targeted spectral emissivity, ensuring that the most effective parameters are identified and utilized.

**ML algorithms**

For both the forward HF and the inverse LF modeling of laser parameters and spectral emissivity in the MF ensemble, the RF algorithm was used.

Additionally, to show that inversely modeling this phenomenon does not yield accurate results, standalone RF, XGB, and LGB were used. More specifically, the RF algorithm is an ensemble learning method used for ML classification and regression. For regression, the algorithm operates by constructing randomly defined decision trees (DTs) at training time and outputting the mean prediction of each individual DT or estimator. Randomness is introduced by selecting a subset of the input features at each split in the training of individual trees, thereby ensuring a de-correlation between the trees and reducing the likelihood of overfitting to the training data. Furthermore, a major advantage of the RF algorithm is that each individual prediction, and not just the aggregate or the mean, can be obtained and used for further analysis. Furthermore, the RF algorithm has the major advantage of being interpretable and is suited for well defined and structured features like the laser processing parameters for the forward mapping of the problem at hand. The RF implementation within the Python ML module scikit-learn 1.2.2. was used in this study[39].

XGB and LGB are both advanced implementations of gradient boosting algorithms. These algorithms build an ensemble of weak prediction models or weak learners (typically DTs), in a sequential manner where each subsequent learner is continuously improved by minimizing the error of previous learners. Both XGB and LGB are designed to be computationally efficient and scalable, capable of handling large datasets and high-dimensional feature spaces like the spectral emissivity problem. XGB's advantage is its efficient optimization and algorithmic enhancements, such as a regularized model to prevent overfitting, and its ability to significantly speed up the training process by using advanced parallel and distributed computing. LGB distinguishes itself with its unique approach to constructing DTs using original techniques such as Gradient-based One-Side Sampling and Exclusive Feature Bundling, which allows large dataset handling by reducing the number of data instances and features without significant loss of accuracy. Finally, LGB grows trees using the leaf-wise technique rather than level-wise as XGB, which often results in faster learning with less memory usage. These specific features make LGB often faster and more resource-efficient than XGB, while XGB may achieve slightly better accuracy given sufficient computational resources since it operates in a way that doesn't include any loss of accuracy. The Python module xgboost 1.7.3 was used for the XGB implementation[40], while the module lightgbm 3.3.5 was used for the LGB implementation[41]. The LGB algorithm does not support regression models with multiple output features thus the scikit-learn 1.2.2 function MultioutputRegressor function was used as a wrapper for the LGB algorithm.

**Data preprocessing and validation strategies**

The total number of experimental samples was 11,759. The spectral emissivities were measured at 822 different wavelength values and can be observed in Figure 1e, while the laser processing parameters, i.e. power, speed and spacing can be observed in Figure 1f. For the forward model, the laser processing parameters were the input features, while for the inverse model, the spectral emissivity values were the inputs. Initially, the experimental dataset was randomly shuffled and split into 8,500 data instances for train and validation (which is 72.3% of the total data), while the rest of the data instances were used as the test set (3,259 or 27.7%). Both the laser processing parameters and the spectral emissivity test set data can be observed in Supplementary Figure 5.

The K-Fold cross-validation strategy was utilized to rigorously investigate the accuracy and robustness (uncertainty of prediction) of both forward and inverse models. This approach involved exclusively employing the train/validation set, which consisted of 8,500 instances. The process entailed randomly selecting a subset of the data for validation, while utilizing the remainder for training. This iteration was repeated $K$ times, with a $K$ value of 10 being applied to both the forward and inverse models, resulting in 10 repetitions of this procedure. Furthermore, the learning curve analysis in conjunction with the K-Fold cross-validation was used to investigate the performance of both inverse and forward models with different sizes of the train and validation datasets i.e. 500, 2500, 5000 and 8500 data instances. The purpose of the learning curve is to assess the minimum number of data instances needed for an accurate and robust model.

Supplementary Figure 8 illustrates the MF ensemble-informed inverse design model's validation strategy, applying a test set of 3,259 instances to assess design novelty and prediction accuracy via NEPD and RMSE metrics, respectively. For each target emissivity, the model suggests multiple laser parameter sets, with the user determining the number based on the RMSE threshold from the HF optimization cycle. These sets are then evaluated against the test set's actual parameters using NEPD and input into the HF forward model to check emissivity predictions against the original targets. Parameters yielding RMSE values under 2% are deemed optimal, confirming the model's efficacy in accurately and reliably predicting laser processing parameters.

For both the forward and inverse models, the PCA was used to compress the spectral emissivity values. The main purpose of the PCA compression in this case was to enhance the computational efficiency of the inverse design model. In Supplementary Figure 4, the results of the PCA compression can be observed. Furthermore, the number of components for the problem at hand was determined by observing if the PCA compression is capable of reconstructing an ideal step function which is usually a target performance for specific applications like TPV emitters. A total of 50 principal components were determined to be sufficient for a reasonable approximation of such an unphysical curve (RMSE = 7.7%), thus the PCA compression applied as a preprocessing step to train both inverse and forward models uses 50 components. The PCA implementation in scikit-learn 1.2.2 was used.

**Optimization goal function and algorithm**

The optimization process in the HF stage of the MF ensemble is based on minimizing the RMSE value (Equation 2) between the forward HF-RF model's predicted spectral emissivity and the target spectral emissivity. More specifically, the objective function is defined as:

$$minimize_x \, RMSE(\epsilon_{target}, \epsilon_P(x)) \qquad (1)$$

subject to
$$x_{lb,1} \leq x_1 \leq x_{ub,1},$$
$$x_{lb,2} \leq x_2 \leq x_{ub,2},$$
$$x_{lb,3} \leq x_3 \leq x_{ub,3}$$

where x = $(x_1, x_2, x_3)^T$ is the optimization design vector in design space $\mathbb{R}^3$, and it represents the three laser processing parameters (power, speed, spacing) used to evaluate the objective function. $\epsilon_{target}$ denotes the target spectral emissivity curve, while $\epsilon_P$ is the forward model predicted spectral emissivity generated by the laser processing parameters. The $x_{lb}$ and $x_{ub}$ are the lower and upper boundaries of the laser processing parameters, which are defined as:

$x_{lb,1}$ = 0.2 W
$x_{ub,1}$ = 1.3 W
$x_{lb,2}$ = 10 mm/s
$x_{ub,2}$ = 700 mm/s
$x_{lb,3}$ = 15 µm
$x_{ub,3}$ = 28 µm

The DE stochastic global optimization algorithm is used to minimize the discrepancy between the predicted and target spectral emissivity defined in Equation 2. DE is a global stochastic optimization algorithm inspired by evolutionary strategies, well-suited for complex problems such as non-linear, non-differentiable, and multi-modal optimization challenges. Thus, it is particularly effective in the inverse design of photonic surfaces with laser processing parameters, a task characterized by its highly multi-modal nature due to numerous one-to-many mappings of spectral emissivity curves to laser parameters. DE iteratively enhances a population of solutions through evolutionary processes including mutation, recombination, and selection. In this context, the Linear Population Size Reduction Success-History Adaptation of Differential Evolution (L-SHADE) variant was employed. L-SHADE leverages adaptive control mechanisms for key hyperparameters, such as the scaling factor and crossover rate, to balance exploration and exploitation within the optimization design space effectively. Other DE hyperparameters used were: initial population size was set 10, external archive size factor was set to 2, historical memory size was set to 6, and the p mutation value was set to 0.11 as recommended in the Python module for numerical optimization indago 0.4.6[42].

**ML model parameter design novelty and accuracy metrics**

For both inverse and forward ML model accuracy assessment, the RMSE and the maximum RMSE were used, as presented in Equation 2 and 3. Respectively. To assess the design novelty of the predicted laser parameters by the inverse model, the NEPD, maximum and average NEPD were used (Equation 4, 5, and 6). NEPD was firstly introduced for this specific purpose[43].

The RMSE measures the error between the model's prediction and the true observations in the experimental dataset. The general model RMSE is defined as:

$$RMSE = \sqrt{\frac{1}{R \times C} \sum_{i=1}^{R} \sum_{j=1}^{C} (y_{i,j}^T - y_{i,j}^P)^2} \qquad (2)$$

In Equation 2, the total number of experimental test samples is denoted as $R$, while $C$ is dependent on type of model being used, more specifically, for the inverse model, $C$ is the number of the laser processing parameters, i.e. 3, while for the forward model, $C$ denotes the number of wavelengths at which the spectral emissivity is measured for all samples, i.e. 822. For assessing the forward model's accuracy, the $y$ values denote the emissivity values at each wavelength, while for the inverse model, $y$ denotes laser parameters. Furthermore, $T$ and $P$ indices in $y_{i,j}^T$ and $y_{i,j}^P$ define the true and predicted values of either laser processing parameters or spectral emissivity.

Subsequently, the maximum RMSE is defined as:

$$Maximum\ RMSE = max\ \{\sqrt{\frac{1}{R}\sum_{j=1}^{R}(y_{i,j}^{T}-y_{i,j}^{P})^{2}}, i=1,...,R\} \tag{3}$$

The maximum RMSE is based on finding the maximum value within a list of RMSE values which correspond to the errors between each model predicted and true experimental instance $i$. This metric was used only for the forward model when spectral emissivity is the output. The maximum RMSE defined in Equation 3 is particularly important since the RMSE (Equation 1) can indicate excellent performance of the model due to a lot of spectral emissivity clustering in a certain range and is not sensitive to the prediction accuracy of the outliers. Finally, the RMSE values for the forward model were additionally normalized by the term $\frac{100}{\epsilon_{max}-\epsilon_{min}}$, where the $\epsilon_{max}$ and $\epsilon_{min}$ are the maximum and minimum theoretical emissivity value, i.e. 1 and 0, respectively. With this normalization, the RMSE for the spectral emissivity obtained with the forward model is expressed as a percentage. The RMSE value for the standalone inverse algorithms was normalized by the range of each of the laser parameters.

The NEPD parameter design novelty metric is defined as:

$$NEPD_i = \frac{1}{\sqrt{3}} \times \sqrt{\sum_{k=1}^{3}(L^{Tn}_{i,k}-L^{Pn}_{i,k})^{2}} \tag{4}$$

$$, where\ L^{Tn}_{i,k} = \frac{L^{T}_{i,k}-L^{T}_{k\ min}}{L^{T}_{k\ max}-L^{T}_{k\ min}}, L^{Pn}_{i,k} = \frac{L^{P}_{i,k}-L^{T}_{k\ min}}{L^{T}_{k\ max}-L^{T}_{k\ min}}$$

$L^{Tn}_{i,k}$ is the normalized $k$-th parameter of the $i$-th true test sample and $L^{Pn}_{i,k}$ is the normalized $k$-th parameter of the predicted test sample. The value $L^{T}_{i,k}$ is the $k$-th parameter of the $i$-th true test sample, and $L^{P}_{i,k}$ is the $k$-th parameter of the $i$-th predicted instance, whereas $L^{T}_{k\ max}$ and $L^{T}_{k\ min}$ are each of the $k$ parameters maximum and minimum values. The parameter index $k$ takes values 1, 2, or 3, indicating the three distinct laser processing parameters being considered for each instance $i$. Additionally, the maximum and average NEPD metrics are defined:

$$\text{Maximum NEPD} = \max\left\{\frac{1}{\sqrt{3}} \times \sqrt{\sum_{k=1}^{3}\left(L^{Tn}_{i,k} - L^{Pn}_{i,k}\right)^2}, i=1,\ldots,R\right\} \quad (5)$$

$$\text{Average NEPD} = \frac{1}{R}\sum_{i=1}^{R}\frac{1}{\sqrt{3}} \times \sqrt{\sum_{k=1}^{3}\left(L^{Tn}_{i,k} - L^{Pn}_{i,k}\right)^2} \quad (6)$$

*ML algorithms hyperparameter optimization*: The hyperparameters for both standalone inverse (RF, LGB, XGB) and forward ML algorithms (HF-RF) were determined using the Python module for hyperparameter optimization Optuna 3.1.0[44]. The goal function for the forward model hyperparameter optimization process was defined as the K-fold cross-validation ($K = 3$) averaged weighted linear combination of the maximum and RMSE values between the model's spectral emissivity prediction and the true spectral emissivity. The weights were set to 0.8 and 0.2 for the maximum RMSE and RMSE, respectively. The reason for this was due to the fact that most of the spectral emissivity curves are clustered near the emissivity values of approximately ~0.25 to ~0.45 (observed in Figure 1e), hence, a low RMSE model score could be deceiving as the main error metric. For the inverse model algorithms, the hyperparameter optimization function was the averaged K-Fold cross-validation RMSE since the laser processing parameters are uniformly distributed in the design space, and not clustered within a certain range. The full hyperparameter values for all three investigated algorithms for both the inverse and forward models are given in Supplementary Table 4, 5, and 6.

**SHAP model feature interpretation algorithm**

SHAP is an advanced technique for interpreting the features of ML models, assigning an importance value to each feature in relation to a specific prediction[36]. Drawing on principles from cooperative game theory, it employs Shapley values to calculate the mean contribution of each feature to the disparity between a given prediction and the baseline—or average—prediction across the dataset.

SHAP values are adept at delineating both the direction (positive or negative) and the magnitude of each feature's impact. They provide a dual perspective on interpretability: offering a detailed understanding of predictions for individual instances (local interpretability), as well as a cumulative view of feature importance across all instances (global interpretability). The SHAP framework is model-agnostic, suitable for application across various ML models, and possesses a distinctive characteristic: the sum of SHAP values equals the difference between the prediction and the dataset's mean prediction, ensuring the fidelity and consistency of the interpretations both locally and globally. Specifically, for tree-based and gradient boosting models, the TreeExplainer function within the Python shap library (version 0.43.0) was employed to interpret a model that predicts spectral emissivity from laser processing parameters, showcasing SHAP's practical utility in providing feature attributions for complex predictive models[45]. The optimized forward HF-RF model and only the train/validation set (8,500 samples with a 75/25% split) were used for SHAP analysis and the separately investigated output features were the average spectral emissivity, and spectral emissivity values at 2.5, 7.25, and 12 μm wavelengths.

## Data availability

All data needed to justify the conclusions in the paper are present in the paper and/or the Supplementary Information. Machine learning models and experimental data are available in https://osf.io/dwgtf/.

## Code availability

The Python code can be accessed at

https://github.com/lukagrbcic/MFEnsemblePhotonicSurfaces

## Acknowledgements

This work was supported by the Laboratory Directed Research and Development Program of Lawrence Berkeley National Laboratory under U.S. Department of Energy Contract No. DE-AC02-05CH11231. This work was also supported by ARPA-E Contract No. 2107-153. J. Mueller was supported by the U.S. Department of Energy, Office of Science, Office of Advanced Scientific Computing Research, Scientific Discovery through Advanced Computing (SciDAC) program through the FASTMath Institute under Contract No. DE-AC36-08GO28308 at the National Renewable Energy Laboratory.

# Inverse design of photonic surfaces on Inconel via multi-fidelity machine learning ensemble framework and high throughput femtosecond laser processing

Luka Grbčić[1, †], Minok Park[2, †], Mahmoud Elzouka[2], Ravi Prasher[2, 3], Juliane Müller[4], Costas P. Grigoropoulos[2, 3], Sean D. Lubner[2, 5, *], Vassilia Zorba[2, 5, *], and Wibe Albert de Jong[1, *]

[1]*Applied Mathematics and Computational Research Division, Computing Science Area, Lawrence Berkeley National Laboratory, Berkeley, California, 94720, USA*

[2]*Energy Storage and Distributed Resources Division, Energy Technologies Area, Lawrence Berkeley National Laboratory, Berkeley, California, 94720, USA*

[3]*Department of Mechanical Engineering, University of California at Berkeley, Berkeley, California, 94709, USA*

[4]*Computational Science Center, National Renewable Energy Laboratory, Golden, Colorado, 80401, USA*

[5]*Department of Mechanical Engineering, Division of Materials Science and Engineering, Boston University, Boston, Massachusetts, 02215, USA*

[*]*Corresponding author(s). E-mail(s): slubner@bu.edu; vzorba@lbl.gov; wadejong@lbl.gov*

[†]*These authors contributed equally to this work.*

**Supplementary Figures and Tables**

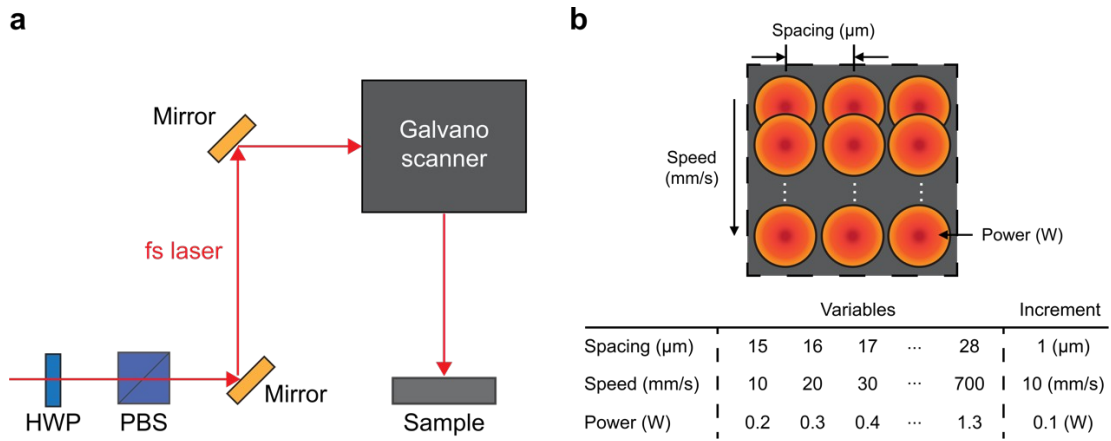

**Supplementary Figure 1.** (a) Schematic representation of the ultrafast fs laser processing setup. Manipulation of laser power was achieved through the utilization of a half wave plate (HWP) and a polarizing beam splitter (PBS). (b) The process involved a raster scanning method to process individual areas under three distinct laser parameters; power, scanning speed, and spacing. Employing these variables as parameter inputs, a total of 11,759 different surfaces were fabricated.

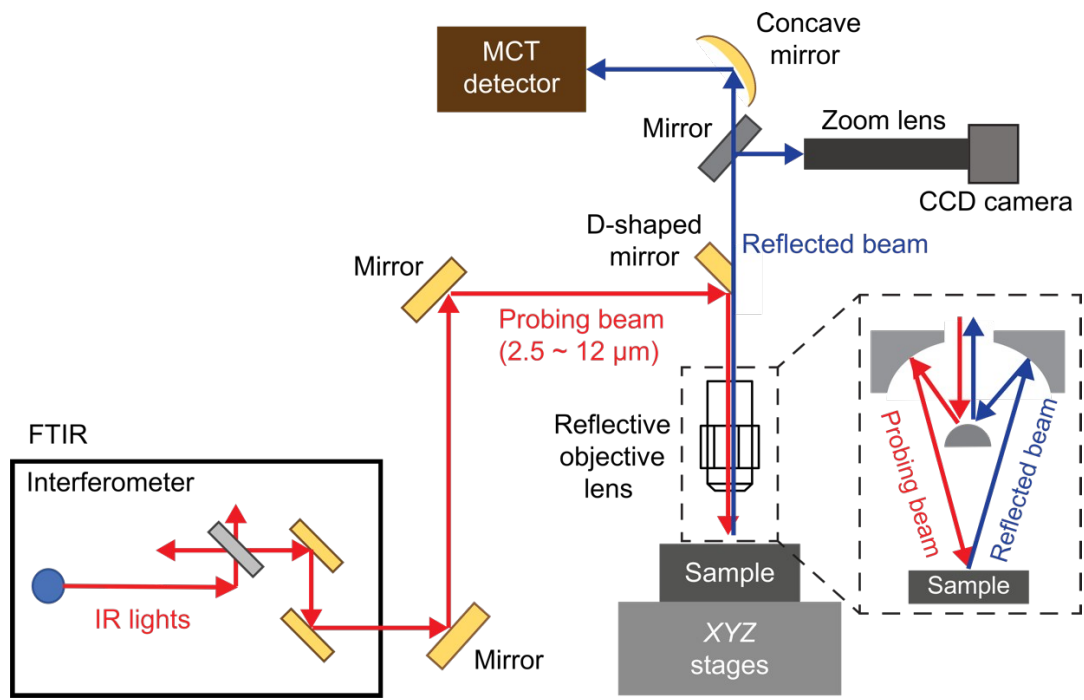

**Supplementary Figure 2.** Schematic of the custom Fourier transform infrared spectrometer (FTIR) microscope system. The infrared (IR) beam is focused on the target surface using a reflective objective lens. To ensure high signal-to-noise ratios for the acquired data, an externally coupled liquid-nitrogen cooled Mercury-Cadmium-Telluride detector is employed in the FTIR system. Synchronization of FTIR equipment and motorized stages enables automated, high-throughput optical property characterization.

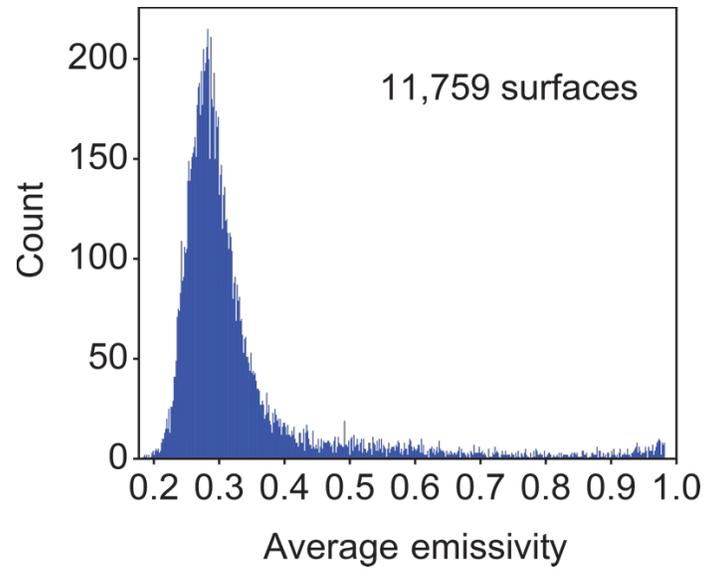

**Supplementary Figure 3.** Average emissivity of 11,759 photonic surfaces fabricated on Inconel using different laser fabrication parameters.

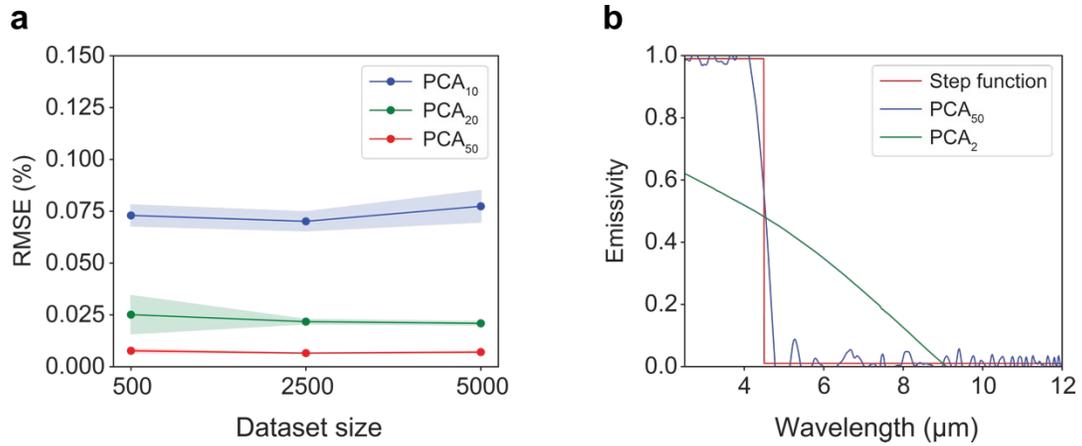

**Supplementary Figure 4.** PCA compression results where the number in the index determines the number of principal components used, i.e. 10 for $PCA_{10}$: (a) Learning curve for a PCA compression-decompression process, for each dataset size, the K-Fold ($K = 10$) cross-validation procedure was used to obtain the uncertainty estimation of the spectral emissivity. (b) Ideal step function approximation with both 2 and 50 component PCA.

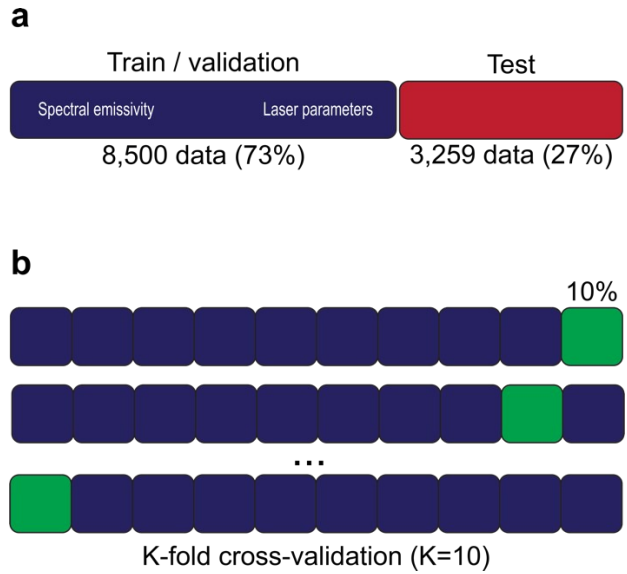

**Supplementary Figure 5.** Data split and validation strategies: (a) The initial experimental data split into 8,500 instances for training and validation, and 3,259 instances for testing of the MF ensemble. (b) The schematic of the K-fold cross-validation procedure–the training set is randomly split into ten different training and validation subsets for model evaluation.

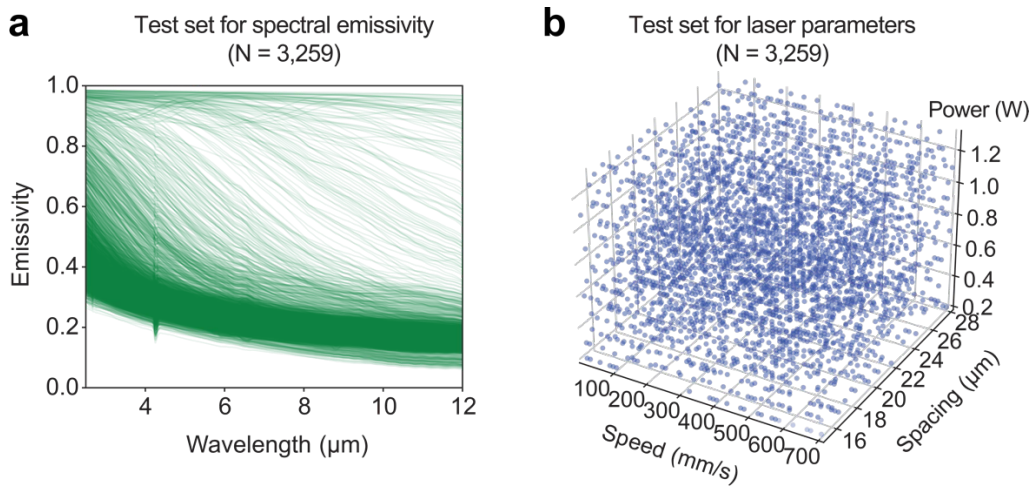

**Supplementary Figure 6.** Experimental test set, where the total number of samples is 3,259 or 27.7% of the total dataset: (a) spectral emissivities, and (b) laser processing parameters.

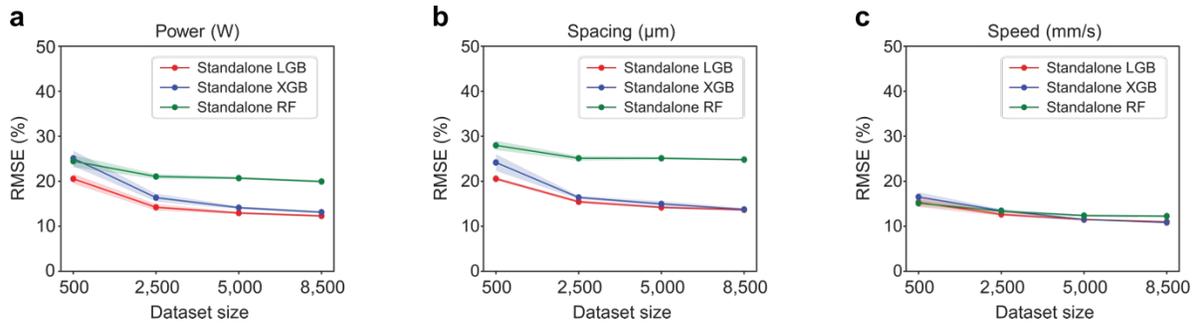

**Supplementary Figure 7.** Learning curves for inverse models including standalone XGB, RF, and LGB with K-Fold cross-validation with $K = 10$ and RMSE for (a) power, (b) spacing, and (c) speed, respectively.

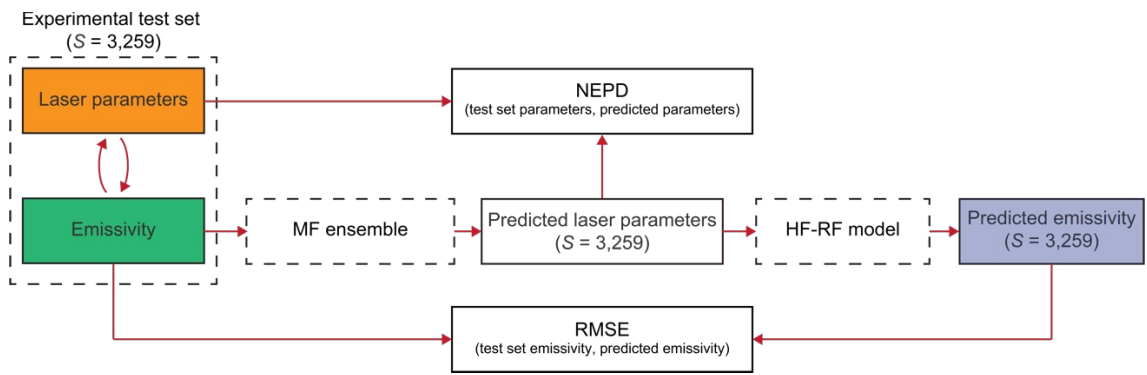

**Supplementary Figure 8.** MF ensemble validation strategy. $S$ is the number of laser parameter sets.

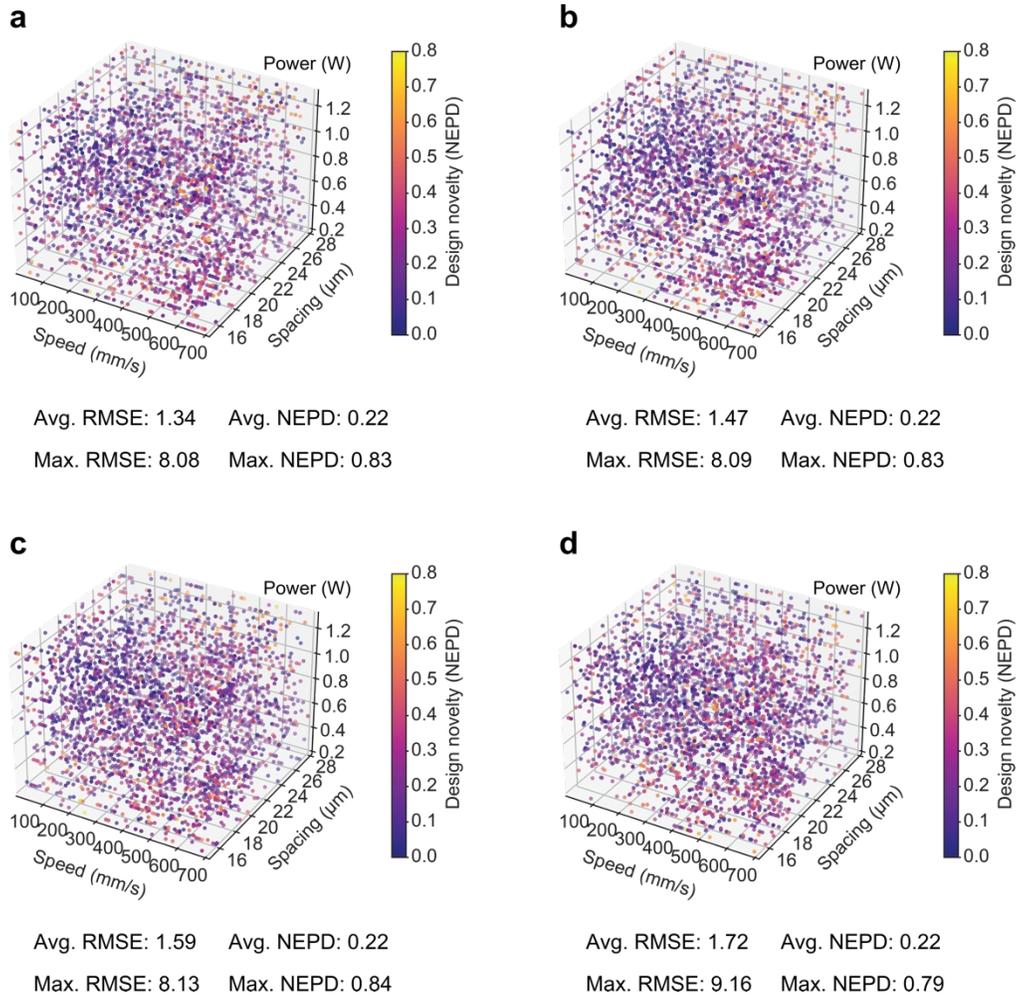

**Supplementary Figure 9.** Results of the MF ensemble for experimental test set spectral emissivities. All laser processing parameters are colored by their NEPD values: (a) Set of the second best performing laser processing parameters in terms of fitness. (b) Set of the third best performing laser processing parameters in terms of fitness. (c) Set of the fourth best performing laser processing parameters in terms of fitness. (d) Set of the fifth best performing laser processing parameters in terms of fitness.

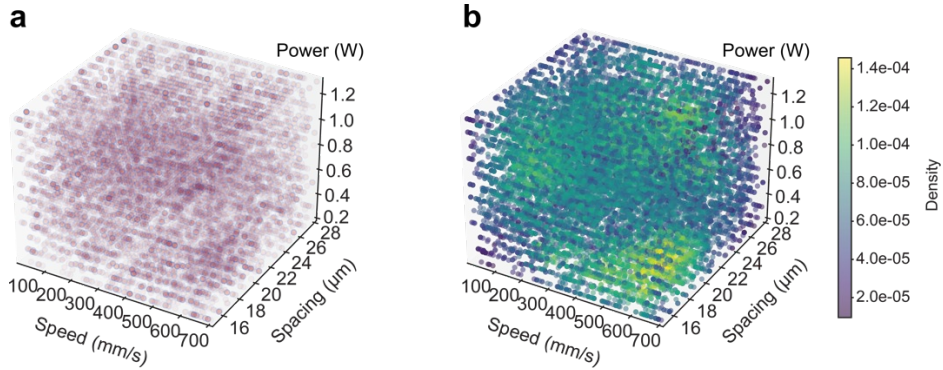

**Supplementary Figure 10.** Visualization of the predicted laser parameters from all 5 sets generated by the MF ensemble: (a) Overlap of all five sets that show where the laser processing parameters cluster the most. (b) Gaussian kernel density estimation plot to indicate where the laser processing parameters from all five prediction sets are statistically concentrated.

The Gaussian kernel density estimation serves as a non-parametric approach for probability density function estimation of a random variable. The plot shows a smooth and continuous representation of the distribution of the laser processing parameters and allows a visualization of areas where the laser processing parameters are most densely concentrated. The *gaussian_kde* function from the Python scipy's 1.11.1 stats module was used[1].

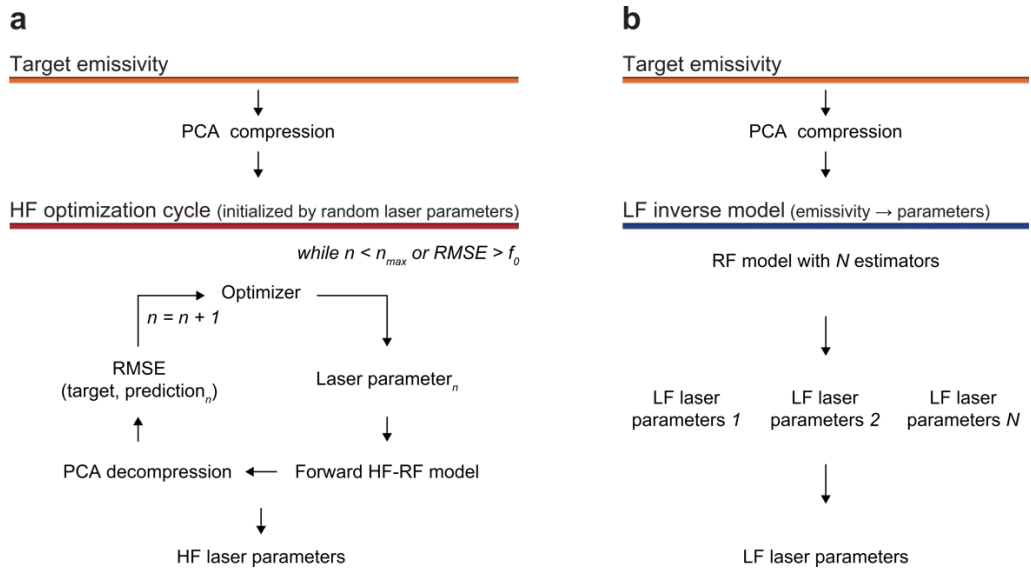

**Supplementary Figure 11.** Two different inverse design model approaches primarily used for comparison with the MF ensemble: (a) Standalone HF inverse design model with no LF initial estimation. (b) Standalone LF inverse design model that does not include any HF optimization cycles.

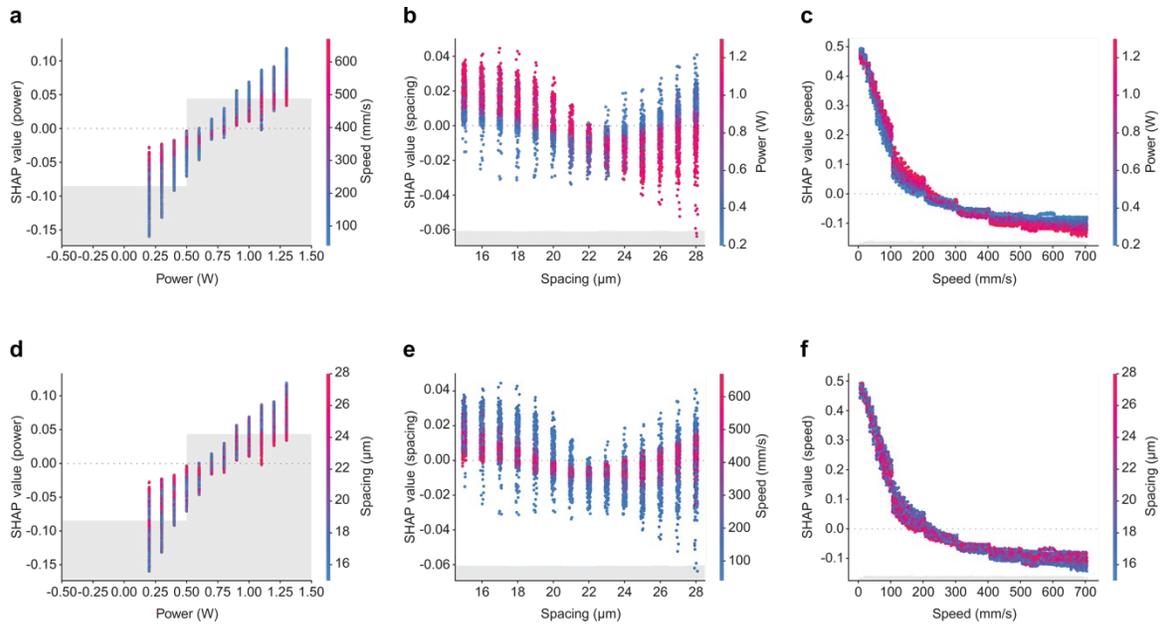

**Supplementary Figure 12.** SHAP features importance analysis of the forward HF-RF model for the emissivity at 2.5 μm wavelength: (a) The relationship between the power feature and the SHAP value colored by the speed feature. (b) The relationship between the spacing feature and the SHAP value colored by the power feature. (c) The relationship between the speed feature and the SHAP value colored by the power feature. (d) The relationship between the power feature and the SHAP value colored by the spacing feature. (e) The relationship between the spacing feature and the SHAP value colored by the speed feature. (f) The relationship between the speed feature and the SHAP value colored by the spacing feature.

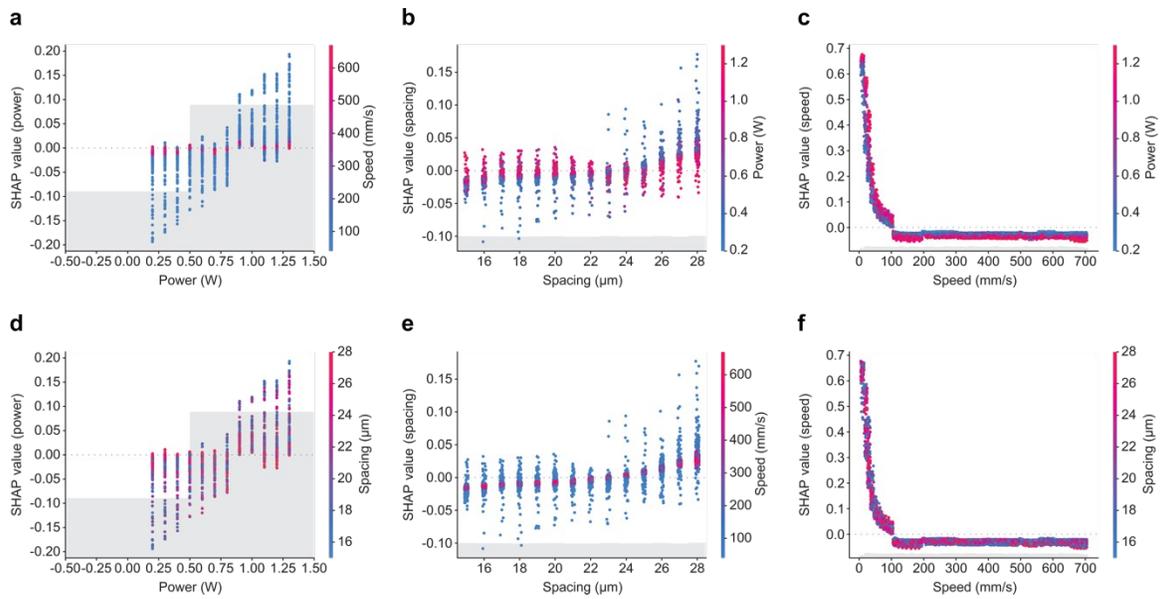

**Supplementary Figure 13.** SHAP features importance analysis of the forward HF-RF model for the emissivity at 7.25 μm wavelength: (a) The relationship between the power feature and the SHAP value colored by the speed feature. (b) The relationship between the spacing feature and the SHAP value colored by the power feature. (c) The relationship between the speed feature and the SHAP value colored by the power feature. (d) The relationship between the power feature and the SHAP value colored by the spacing feature. (e) The relationship between the spacing feature and the SHAP value colored by the speed feature. (f) The relationship between the speed feature and the SHAP value colored by the spacing feature.

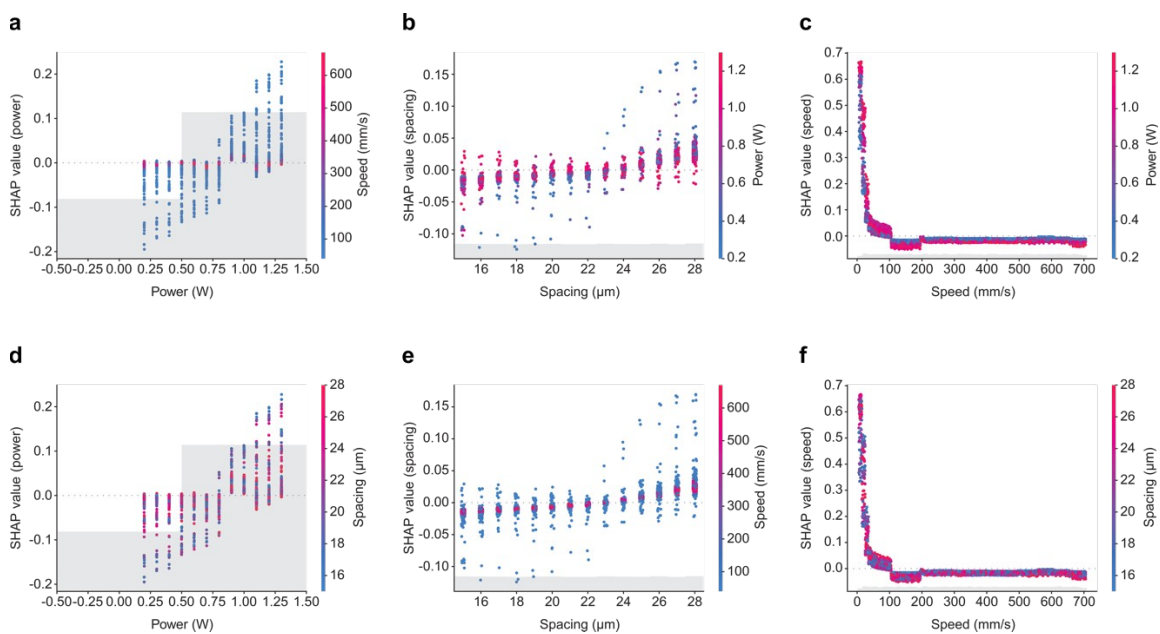

**Supplementary Figure 14.** SHAP features importance analysis of the forward HF-RF model for the emissivity at 12 μm wavelength: (a) The relationship between the power feature and the SHAP value colored by the speed feature. (b) The relationship between the spacing feature and the SHAP value colored by the power feature. (c) The relationship between the speed feature and the SHAP value colored by the power feature. (d) The relationship between the power feature and the SHAP value colored by the spacing feature. (e) The relationship between the spacing feature and the SHAP value colored by the speed feature. (f) The relationship between the speed feature and the SHAP value colored by the spacing feature.

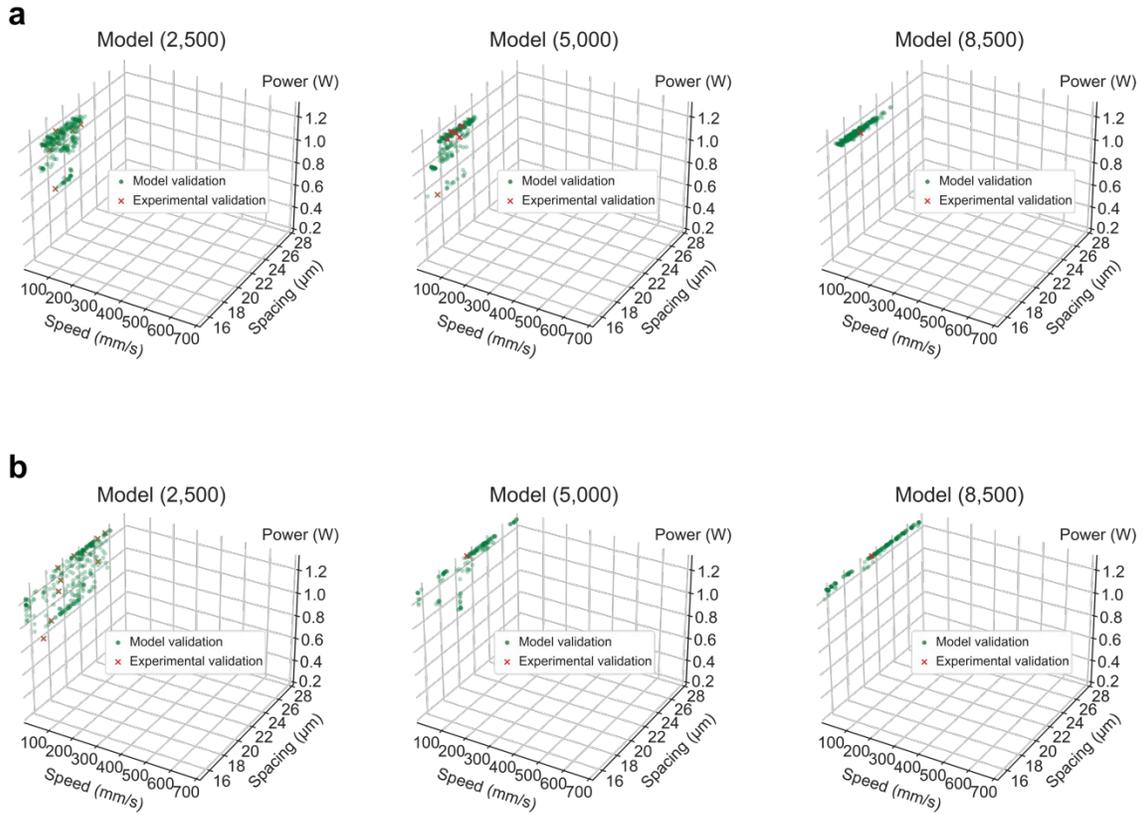

**Supplementary Figure 15.** Inverse designed laser parameters. Laser parameters for (a) lead-selenide thermophotovoltaic emitter, and (b) blackbody surface using MF ensemble trained by 2,500, 5,000, and 8,500 datasets.

| Power (W) | 0.2 | 0.3 | 0.4 | 0.5 | 0.6 | 0.7 | 0.8 | 0.9 | 1.0 | 1.1 | 1.2 | 1.3 |
|---|---|---|---|---|---|---|---|---|---|---|---|---|
| Fluence (J/cm$^2$) | 0.28 | 0.42 | 0.57 | 0.70 | 0.85 | 0.99 | 1.13 | 1.27 | 1.41 | 1.56 | 1.70 | 1.84 |
| Intensity (W/cm$^2$) | $0.28 \times 10^5$ | $0.42 \times 10^5$ | $0.57 \times 10^5$ | $0.70 \times 10^5$ | $0.85 \times 10^5$ | $0.99 \times 10^5$ | $1.13 \times 10^5$ | $1.27 \times 10^5$ | $1.41 \times 10^5$ | $1.56 \times 10^5$ | $1.70 \times 10^5$ | $1.84 \times 10^5$ |

**Supplementary Table 1.** Laser processing parameters in power (W), intensity (W/cm$^2$), and pulse fluence (J/cm$^2$) at a 100 kHz repetition rate.

| Model | Average time (ms) | Standard deviation (ms) |
|:---:|:---:|:---:|
| MF ensemble | 620 | 60 |
| Standalone HF model | 780 | 616 |
| Standalone LF model | 1.53 | 0.82 |

**Supplementary Table 2.** Average inference time for 1000 test set predictions. The $n_{max}$ hyperparameter was set to 100 and 25 for standalone HF model and MF ensemble, respectively, while $f_0$ is set to 2% for both. For the standalone HF model and MF ensemble to be comparable in terms of accuracy, the large discrepancy in the $n_{max}$ value is needed. The experiment is run on a Lenovo Thinkpad X1 ExtremeGen 4 laptop with 11th Gen Intel Core i7-11800H (2.30 GHz) and 16 GB RAM.

| Target/Dataset size | Total predicted parameters | Percentage of novel parameters (Model validation) | Percentage of novel parameters (Experimental validation) | Percentage of predicted parameters used during training |
|---|---|---|---|---|
| TPV emitter/2,500 | 339 | 93,80% | 2,95% | 3,25% |
| TPV emitter/5,000 | 322 | 93,17% | 1,86% | 4,97% |
| TPV emitter/8,500 | 475 | 97,48% | 0,21% | 2,31% |
| Near-perfect emitter/2,500 | 311 | 91,97% | 3,53% | 4,50% |
| Near-perfect emitter/5,000 | 175 | 90,86% | 0,57% | 8,57% |
| Near-perfect emitter/8,500 | 228 | 92,98% | 0,44% | 6,58% |

**Supplementary Table 3.** MF ensemble 100 run statistics for the lead-selenide TPV and near-perfect emitter spectral emissivity targets, and all considered training dataset sizes (2,500, 5,000 and 8,500, respectively). The total predicted parameters are the number of laser parameter set predictions after duplicate filtering.

| LGB hyperparameter | Standalone inverse model |
|---|---|
| boosting_type | gbdt |
| num_leaves | 26 |
| feature_fraction | 0.95 |
| bagging_fraction | 0.94 |
| bagging_frequency | 1 |
| min_child_samples | 7 |
| learning_rate | 0.09 |
| n_estimators | 100 |

**Supplementary Table 4.** Standalone inverse model optimized hyperparameters for the LGB algorithm.

| XGB hyperparameter | Standalone inverse model |
|:---:|:---:|
| max_depth | 8 |
| n_estimators | 100 |
| learning_rate | 0.07 |
| reg_lambda | 5.1 |
| reg_lambda | 4.81 |

**Supplementary Table 5.** Standalone inverse model optimized hyperparameters for the XGB algorithm.

| RF hyperparameter | Standalone inverse model | Forward HF-RF model |
|---|---|---|
| max_depth | 10 | 10 |
| n_estimators | 450 | 450 |
| min_samples_leaf | 1 | 1 |
| max_features | sqrt | auto |

**Supplementary Table 6** Forward HF-RF model and standalone inverse optimized hyperparameters for the RF algorithm.